\pdfoutput=1
\documentclass[11pt]{article}
\usepackage[dvipsnames]{xcolor}
\usepackage{emnlp2021}
\usepackage{times}
\usepackage{latexsym}
\usepackage[T1]{fontenc}
\usepackage[utf8]{inputenc}
\usepackage{microtype}
\usepackage{todonotes}
\usepackage{booktabs}
\usepackage{amsmath}
\usepackage{multicol}
\usepackage{multirow}
\usepackage{comment}
\usepackage{url}

\newcommand{\citepos}[1]{\citeauthor{#1}'s (\citeyear{#1})}
\newcommand{\en}{\textsc{en}}
\newcommand{\fakeen}{\textsc{[en]}}
\newcommand{\xx}[1]{\textsc{#1}}

\newcommand{\enxx}[1]{\textsc{en$\sim$#1}}
\newcommand{\fakeenxx}[1]{\textsc{[en$\sim$#1]}}
\newcommand{\orig}{\en{} + \fakeen{}}
\newcommand{\biling}[1]{\en{} + \fakeenxx{#1}}

\title{The Impact of Positional Encodings on Multilingual Compression}

\author{Vinit Ravishankar \\
  Language Technology Group \\
  University of Oslo \\
  Norway \\
  \texttt{vinitr@ifi.uio.no} \\\And
  Anders Søgaard \\
  Department of Computer Science \\
  University of Copenhagen \\
  Denmark \\
  \texttt{soegaard@di.ku.dk} \\}

\begin{document}
\maketitle
\begin{abstract}
In order to preserve word-order information in a non-autoregressive setting, transformer architectures tend to include positional knowledge, by (for instance) adding positional encodings to token embeddings. Several modifications have been proposed over the sinusoidal positional encodings used in the original transformer architecture; these include, for instance, separating position encodings and token embeddings, or directly modifying attention weights based on the distance between word pairs. We first show that surprisingly, while these modifications tend to improve monolingual language models, none of them result in better multilingual language models. We then answer why that is: Sinusoidal encodings were explicitly designed to facilitate compositionality by allowing linear projections over arbitrary time steps. Higher variances in multilingual training distributions requires higher compression, in which case, compositionality becomes indispensable. Learned absolute positional encodings (e.g., in mBERT) tend to approximate sinusoidal embeddings in multilingual settings, but more complex positional encoding architectures lack the inductive bias to effectively learn compositionality {\em and} cross-lingual alignment. In other words, while sinusoidal positional encodings were originally designed for monolingual applications, they are particularly useful in multilingual language models.
\end{abstract}

\section{Introduction}
Multiple recent papers have attempted to pinpoint precisely what components of multilingual language models enable cross-lingual transfer. \citet{pires_HowMultilingual_2019} show that although wordpiece overlap tends to improve cross-lingual transfer performance, even languages with different scripts (and no shared subwords) may enable zero-shot transfer.~\citet{wu_betobentz_2019} report similar results on a wider range of tasks.
\citet{artetxe_CrosslingualTransferability_2020} show that neither a shared vocabulary nor joint multilingual pre-training are necessary to train successful multilingual models. \citet{k_CrossLingualAbility_2020} find that model depth is a contributor to transfer performance, but that reducing the number of self-attention heads does not have much of an effect. 

Our starting point is~\citet{dufterIdentifyingNecessaryElements2021}, who claim that a) multilingual compression is caused by forced parameter sharing across languages, and that b) positional encodings play a significant role in the creation of a multilingual space, even in the absence of shared subwords and shared special tokens, like delimiters.

\begin{table}[]
\scriptsize 
    \centering
    \begin{tabular}{lll}
         \toprule 
         {\bf Sinusoidal}&See \S2&\citet{vaswani_attention_2017}  \\
         {\bf Absolute}&$((w_i+p_i)W^{Q,1})$&\citet{devlin_bertpretraining_2019}\\
         &$((w_j+p_j)W^{K,1})^{\top}$&\\
        {\bf TUPE}&$(x_i^lW^{Q,l})(x_j^lW^{K,l})^{\top} +$&\citet{keRethinkingPositionalEncoding2020}\\&$ (p_iU^Q)(p_jU^K)^{\top}$\\
        {\bf TUPE}(r)&\ldots $+b_{j-i}$ & \\
         {\bf Relative}(k)&$(x_iW^Q)(x_jW^K + a_{ij})^{\top}$&\citet{shawSelfAttentionRelativePosition2018}\\ 
         {\bf Relative}(k/q)&$(x_iW^Q + a_{ij})$&\citet{huangImproveTransformerModels2020}\\&$(x_jW^K + a_{ij})^{\top}$&\\ 
         \bottomrule 
    \end{tabular}
    \caption{We compare six positional encodings and their impact on cross-lingual generalization in multilingual language models}
    \label{tab:embeds}
\end{table}

\paragraph{Contributions} We build on \citet{dufterIdentifyingNecessaryElements2021} and demonstrate, through a series of experiments on synthetic and real data, that the choice of positional encoding mechanism has a significant effect on cross-lingual model performance: While many positional encodings have been proposed in monolingual settings as improvements over sinusoidal or absolute positional encodings, originally proposed in \citet{vaswani_attention_2017} and \citet{devlin_bertpretraining_2019}, including untied positional encodings (TUPE; \citet{keRethinkingPositionalEncoding2020}) and relative positional encodings \cite{shawSelfAttentionRelativePosition2018,huangImproveTransformerModels2020}, none of these better facilitate cross-lingual compression or sharing. In fact, multilingual language models trained with untied or relative positional encodings exhibit {\em much worse} cross-lingual performance. We show that this is because sinusoidal embeddings facilitate compositionality, which we argue is particularly important for cross-lingual compression. We present a method for quantifying the compositionality of positional encodings, and find additional evidence for this hypothesis in word-position correlations and ablation studies. We are, to the best of our knowledge, the first to show this asymmetry between monolingual and multilingual language model training. Our experiments rely on the protocols in \citet{dufterIdentifyingNecessaryElements2021}, but in addition to simple experiments with their Bible data, we also replicate all our experiments on Wikipedia data. Rather than relying on deterministic perturbations of data, as in \citet{dufterIdentifyingNecessaryElements2021} and \citet{sinha_maskedlanguage_2021}, we make novel use of Galactic Dependencies \cite{wang_galacticdependencies_2016} in our experiments. Based on our experiments, we recommend caution when adopting methods developed for monolingual language models when training multilingual models, as well as that future work on positional encoding mechanisms also provides evaluations in multilingual settings.

\section{Positional encodings} 
\label{sec:pos}
\label{ssec:pos_bg}
Positional encodings have been a mainstay of non-autoregressive transformer-based models right since~\citet{vaswani_attention_2017} first proposed the transformer architecture. The motivation being that given that transformers\footnote{Note that we use "transformers" as shorthand for transformer encoders used for masked language modelling.} are order-invariant (as opposed recurrent or convolutional networks), there must be some injection of word order into the encoder.
Rather than using conventional "embeddings", \citet{vaswani_attention_2017} use fixed {\bf sinusoidal} position encodings, where each dimension characterises a sinusoidal waveform of a fixed frequency.  Specifically, each encoding $p$ is given as: 
    \begin{align*}
        p_{(pos,2i)} &= sin(pos/10000^{2i/d_{\mathrm{model}}})\\
        p_{(pos,2i+1)} &= cos(pos/10000^{2i/d_{\mathrm{model}}})
    \end{align*}
    where $pos$ is the position and $i$ is the dimension. They add these encodings to token representations before passing the sum to the first layer of the self-attention mechanism.

Several alternatives to sinusoidal encodings have been proposed since \citet{vaswani_attention_2017}. Most multilingual models tend to use BERT-style~\citep{devlin_bertpretraining_2019} learnt {\bf absolute} positional encodings, where a unique vector is learned and assigned to each position; these vectors are then added to word representations before being passed to the self-attention mechanism. 


As an alternative to such position representations, where every position is represented by a unique vector, \textbf{relative} positional encodings have been proposed~\citep{shawSelfAttentionRelativePosition2018,huangImproveTransformerModels2020}. Rather than assigning representations to tokens based on their position, relative positional encoding involves assigning representations to position-position pairs; typically, these encodings are calculated separately and added to the attention matrix. We evaluate both the encodings proposed in \citet{shawSelfAttentionRelativePosition2018} and the encodings proposed in \cite{huangImproveTransformerModels2020} in our experiments below. 


\citet{he_debertadecodingenhanced_2021} propose eliminating position-position correlations, and using separate parameters for word and position representations; \citet{wang_selfattentionstructural_2019a} propose using dependency trees instead of raw sequential positions. \citet{keRethinkingPositionalEncoding2020} recommend eliminating the addition operation in BERT-style representations; they argue that word-position correlations are effectively nil, and that the addition introduces unnecessary noise. We evaluate two {\bf untied} positional encodings proposed in \citet{keRethinkingPositionalEncoding2020} (TUPE). TUPE modifies absolute representations by a) untying word-position correlations; b) using a separate set of parameters for positional attention and c) untying \texttt{[CLS]} tokens from positions.

We refer to recent surveys~\citep{dufterPositionInformationTransformers2021,wangPOSITIONEMBEDDINGSBERT2021} for a more detailed treatment of position encoding methods. We provide a summary of our methods in Table~\ref{tab:embeds}. $W^{Q,l}$ and $W^{K,l}$ represent the query/key weights for the attention mechanism at some layer $l$, and $a_{ij}$ or $b_{j-i}$ are learnt vectors corresponding to the offset $j - i$. Note that the untied position-position term $(p_iU^Q)(p_jU^K)^\top{}$ is added at every layer.

The above positional encodings have been introduced in the context of monolingual pretrained language models, and there has been only a limited amount of work addressing the effect of positional encodings on multilingual models.
\citet{liu_improvingzeroshot_2020} find that positional information tends to hurt machine translation, as the encoder learns a word-order bias towards the source languages.\footnote{The results in \citet{liu_improvingzeroshot_2020} apply to zero-shot generalization of fine-tuned, task-specific models and not to how multilingual language models are pretrained. In their experiments, they rely on a pretrained language model with absolute positional encodings. In fact, what they show is that freezing these during fine-tuning helps cross-lingual zero-shot generalization.}
\citet{artetxe_CrosslingualTransferability_2020} find that language-specific positional representations help in an adapter-based training scenario.
\citet{ding_selfattentioncrosslingual_2020} attempt to account for structural differences between languages by using bracketing transduction grammar trees to reorder position labels (and find that it helps). 
\citet{liu_importanceword_2020} find that models that are relatively agnostic to word-order tend to perform better in cross-lingual settings; they hypothesise that large multilingual encoders, being trained on languages with drastic differences in word orders, tend to have order-agnostic positional encodings, and thus discourage fine-tuning positional encodings downstream. Contemporaneous with this work,~\citet{sinha_maskedlanguage_2021} show that positional information is important for monolingual models even given unnatural, randomly shuffled word ordering. 

\citet{dufterIdentifyingNecessaryElements2021} present a set of experiments training smaller language models on bilingual corpora, consisting of the same corpus in English and "fake-English", which is English with a shifted BPE vocabulary. They evaluate retrieval and translation scores at different layers; gold alignments are easy to derive given that the corpora are effectively parallel corpora, and that the vocabularies for both halves are effectively the same. As we build on these experiments, we adopt slightly simplified notation, and denote vocabulary-shifted corpora with square brackets, eg. \fakeen{}. 

   
   \begin{figure*}[h]
    \centering
    \includegraphics[width=\textwidth]{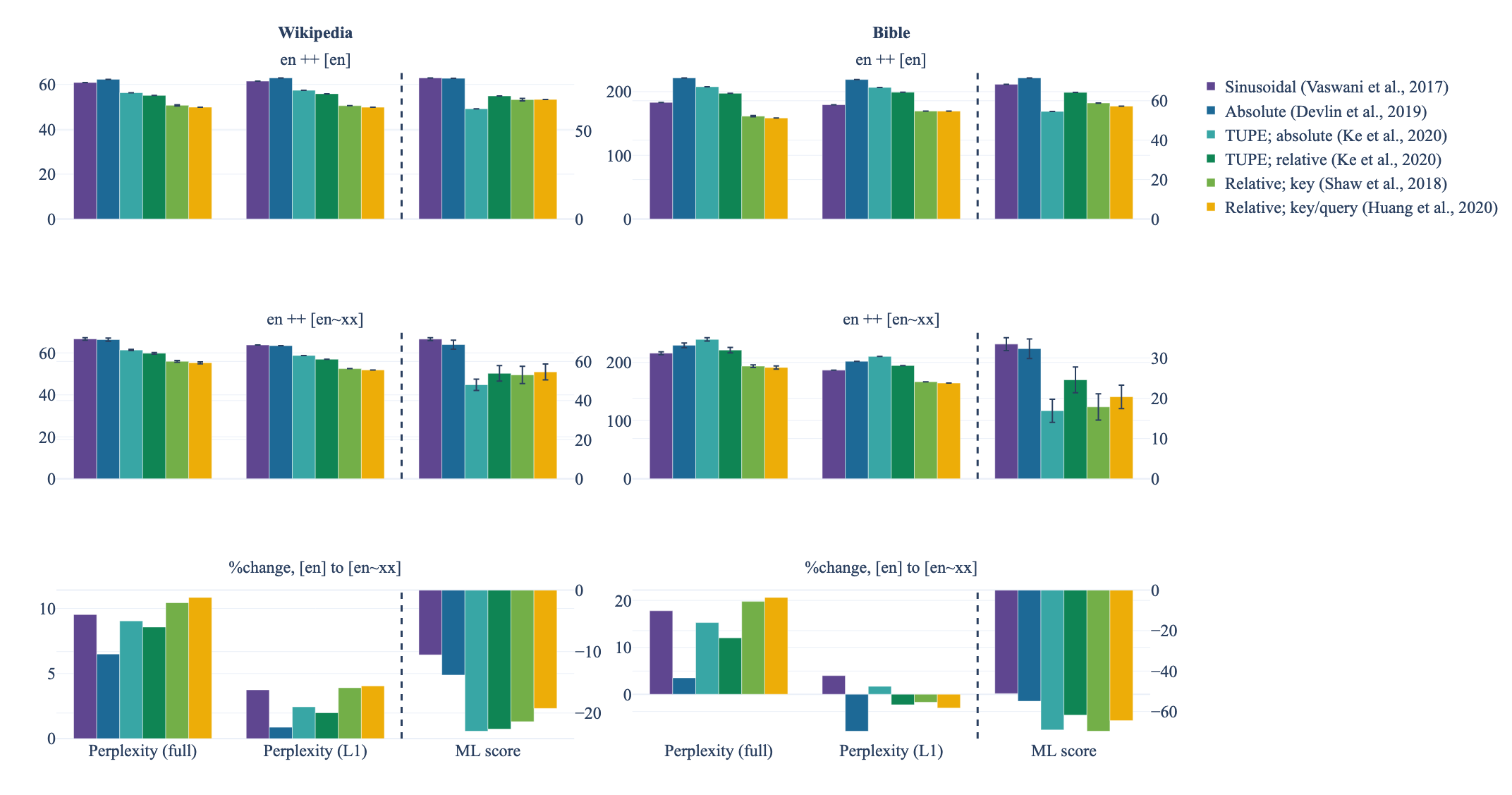}
    \caption{{\bf Main results:} While untied and relative positional encodings are superior to sinusoidal and absolute positional encodings in the monolingual setting, they are clearly worse in the multilingual setting, especially for structurally different languages. The multilingual (ML) scores are computed as in \citet{dufterIdentifyingNecessaryElements2021}. Note also that results are averages across seven different word orders (see \S3).}
    \label{fig:macro}
\end{figure*}

\section{Experiments}
\label{sec:exp}

\paragraph{Galactic Dependencies}
\label{ssec:exp_struct}
A drawback of the multilingual experiments presented in~\citet{dufterIdentifyingNecessaryElements2021} is that \en{} and \fakeen{} effectively have the same structure. While the authors attempt to control for this in additional experiment where word order in \fakeen{} is completely reversed, this does not resemble realistic differences across languages. Using true multilingual corpora is, however, difficult: our retrieval and translation tasks are easy to bootstrap precisely because we have faux-parallel corpora, with effectively pre-aligned vocabulary. 

To induce structural diversity in our corpora, therefore, we reorder our corpora using Galactic Dependencies (GD) models~\citep{wang_galacticdependencies_2016}. Briefly, GD models sample ordering statistics based on dependency relations for the dependants of verbs and/or nouns from some superstrate language~\xx{xx}; when applied to sentences in some substrate language (in the context of our experiments, \en{}), the models reorder dependants of \textsc{verb} and/or \textsc{noun} nodes to match the ordering statistics of the substrate language they were trained on. We opt to reorder both nominal and verbal arguments, and follow the authors in denoting the sampling operation with a $\sim$, giving us for eg.~\enxx{xx} for an English language corpus, with dependent order statistics adapted from some language \xx{xx}. Table~\ref{tab:gd} contains an example sentence and some of its reorderings. 

Note that GD reordering only works for projective sentences, and rather than retain un-reordered non-projective sentences, we exclude them from all our corpora.

\begin{table}[h]
    \centering
    \resizebox{0.5\textwidth}{!}{\begin{tabular}{cc}
        \midrule
        \xx{en} & So there were fourteen generations from Abraham to David .\\
        \enxx{ar} & . there were So generations fourteen from Abraham to David\\
        \enxx{de} & there were So from Abraham to David fourteen generations .\\
        \enxx{eu} & there were So David to Abraham from generations fourteen .\\
        \enxx{fi} & Abraham from David to fourteen generations there were So .\\
        \enxx{fr} & fourteen generations from Abraham to David were there So .\\ 
        \enxx{hi} & there So David to Abraham from fourteen generations were .\\
        \enxx{sv} & there were So generations from Abraham to David fourteen .\\
        
        \bottomrule
    \end{tabular}}
    \caption{An example sentence from the easy-to-read Bible with its GD reorderings.}
    \label{tab:gd}
\end{table}

This approach, while simple and useful, does have several limitations. Predominantly, because our reordering is fundamentally syntactic/structural, our fake languages still maintain both the morphology of the source language (English in our case), and the same vocabulary distribution. Thus, although scrambling ought to affect context and neighbourhoods, an English token and its corresponding fake token have exactly the same unigram distribution.

\paragraph{Training}
\label{ssec:exp_training}
Our model of choice is an underparameterised BERT, as in \citet{dufterIdentifyingNecessaryElements2021}. We train multiple such underparameterised BERT models, each with a different encoding mechanism from Section~\ref{ssec:pos_bg}, on two bilingual corpora:

\begin{description}
    \item[\orig{}] - a bilingual corpus comprised of English, and a fake vocab-shifted English. 
    \item[\biling{xx}] - a bilingual corpus comprised of English, and a fake English that has had its constituents reordered to match the distribution of some language \xx{xx}.
\end{description}

We reorder our English starting point according to seven different faux-languages (just "languages" for brevity): Arabic, German, Basque, Finnish, French, Hindi and Swedish. Note that given that our starting point was English, there was no way for us to control for morphological differences; as such, languages with freer word order (like Basque) are likelier to make our English corpora ambiguous. 

We use two corpora in this work: the first is the Bible splits from~\citet{dufterIdentifyingNecessaryElements2021}, with the English easy-to-read Bible as the training split, and the KJV Bible as validation. The second corpus uses the English Wikipedia as the training split, and Common Crawl as validation. We present corpus statistics in Table~\ref{tab:corpus}. For each corpus, we learn and apply a BPE vocabulary of size 2048.

\begin{table}[h]
    \centering
    \begin{tabular}{ccc}
        & Train & Validation  \\
        \midrule
        Bible & 30602 & 9080 \\
        Wikipedia & 50000 & 20000 \\
        \bottomrule
    \end{tabular}
    \caption{Corpus sizes in sentences (two languages per corpus)}
    \label{tab:corpus}
\end{table}

Following \citet{dufterIdentifyingNecessaryElements2021}, our BERT models all have a single head and 12 layers. We reduce the dimensionality of the  encoder layers to 64, and the feed-forward layers to 256. Each model is trained for 100 epochs with three different random seeds (0, 42 and 100), giving us a total of 7 languages x 6 encoding methods x 3 seeds x 2 corpora = 252 models. We implement our code\footnote{\url{github.uio.no/vinitr/multilingual-position}} in the transformers library~\citep{wolf-etal-2020-transformers}. For learned absolute and the two relative encoding models, we use the default implementations, that scale attention operations by a scaling factor of $\frac{1}{\sqrt{d}}$. For our untied models, we adjust our scaling factor to $\frac{1}{\sqrt{2d}}$ as in the original paper~\citep{keRethinkingPositionalEncoding2020}. For sinusoidal representations, while \citet{vaswani_attention_2017} multiply token embeddings by $\sqrt{d}$ to avoid drowning them out with the $[-1, 1]$ sinusoidal encoding range, we find that our default embedding size is too small for this to have an effect, and instead scale up token embeddings by $2\sqrt{d}$ before adding positional encodings.

For all parameterised encoding models except TUPE (relative), we use a maximum of $k = 512$ positions; the concrete transformers implementation of the relative methods means that this gives us 1023 total offsets. \footnote{In line with~\citet{shawSelfAttentionRelativePosition2018}, we also attempted to use $k = 16$ for the relative key model, but saw no difference in results.} For TUPE (relative), we use a maximum of $k = 128$ positions, divided into 32 bins with logarithmically increasing bin sizes; this is taken from the original implementation in~\citet{keRethinkingPositionalEncoding2020}.

\section{Evaluation}
We adopt~\citepos{dufterIdentifyingNecessaryElements2021} evaluation pipeline, evaluating each of our models at layers 0 and 8; we also describe a multilingual score, which is defined as the average accuracy for the retrieval and translation tasks, at layers 0 and 8. We also measure perplexity, both on the monolingual first half of the corpus, and on both halves combined. Note that true perplexities for masked language models are intractable~\citep{wang2019bert, Salazar_2020}. We use a trivial approximation and calculate perplexity based on the prediction loss for each masked token; note that while these suffice for comparison purposes, they are not true perplexities and should not be taken as such outside the context of these experiments.

\begin{table*}
\centering
\resizebox{\textwidth}{!}{\begin{tabular}{c|ccccccc|ccccccc}
&\multicolumn{7}{c|}{Wiki/CC} & \multicolumn{7}{c}{Bible} \\
\multirow{2}{*}{Embedding} & \multicolumn{2}{c}{Perplexity} & \multicolumn{2}{c}{Retrieval} & \multicolumn{2}{c}{Translation} & \multirow{2}{*}{ML score} & \multicolumn{2}{c}{Perplexity} & \multicolumn{2}{c}{Retrieval} & \multicolumn{2}{c}{Translation} & \multirow{2}{*}{ML score} \\
& Full & L1 & 0 & 8 & 0 & 8 & & Full & L1 & 0 & 8 & 0 & 8\\
\midrule
Sinusoidal &66.73 &63.96 &37.43 &\textbf{97.29} &\textbf{77.03} &\textbf{64.07} &\textbf{68.95} &215.77 &186.53 &4.82 &\textbf{54.09} &47.09 &\textbf{27.62} &\textbf{33.4} \\
Absolute &66.35 &63.44 &\textbf{52.35} &96.53 &76.05 &53.62 &68.59 &229.28 &201.88 &\textbf{9.62} &52.51 &\textbf{47.6} &19.36 &32.27 \\
TUPE (absolute) &61.4 &58.77 &9.61 &84.72 &65.89 &36.57 &48.07 &239.48 &210.16 &1.65 &28.86 &28.85 &8.12 &16.87 \\
TUPE (relative) &59.81 &56.96 &16.25 &88.5 &71.7 &40.54 &53.89 &221.04 &194.58 &2.54 &41.34 &40.39 &13.92 &24.55 \\
Relative (key) &55.98 &52.5 &20.2 &87.36 &73.09 &31.38 &53.09 &193.49 &166.75 &2.18 &28.46 &30.43 &10.25 &17.83 \\
Relative (key/query) &\textbf{55.28} &\textbf{51.83} &21.24 &88.04 &73.58 &34.42 &54.64 &\textbf{191.23} &\textbf{164.41} &2.4 &31.6 &34.26 &13.03 &20.32 \\
\end{tabular}}
\caption{Detailed results, averaged across our faux-languages. Best results per metric in bold.}
\end{table*}
We present our results (averaged out over faux-languages) in Figure~\ref{fig:macro}, with full results in Appendix~\ref{sec:app_full}. As expected, the more recent positional encodings are superior to sinusoidal or absolute positional encodings in the monolingual setting; but somewhat surprisingly, sinusoidal and absolute positional encodings are clearly outperforming the more recent approaches in the multilingual setting. 
We also note that the gap in multilingual performance only grows larger when a different word order is imposed on the target language; see the bottom row of Figure~\ref{fig:macro}. Interestingly, switching to structurally different L2s can sometimes reduce the language modelling perplexity of the L1: this could be due to regularisation induced by structural differences.

\paragraph{Typological differences}
\begin{figure}[t]
    \centering
    \includegraphics[width=0.5\textwidth]{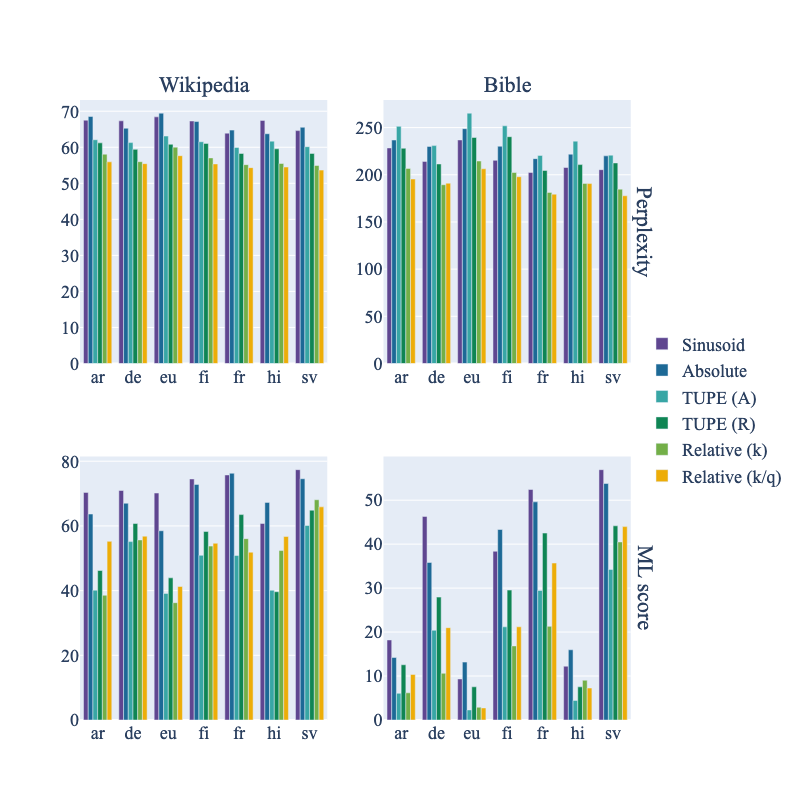}
    \caption{(Full) perplexity and ML score across languages.}
    \label{fig:xling}
\end{figure}
We discuss "typology" with a caveat: our experiments with GD only alter word order, which means that all our altered-structure experiments still have English morphology. As such, it is impossible to talk about non-English languages; only about non-English word-order tendencies, when induced in English. Having said that, when we measure performance variation across languages (Figure~\ref{fig:xling}), our results are more or less what one would expect: performance is decent for relatively rigid word-order languages, and poorer for languages that have complex morphology. 

Interestingly, SVO languages consistently tend to perform better than our three non-SVO languages (Basque, Hindi and Arabic); this could be due to VSO/SOV languages requiring morphology to disambiguate between adjacent nominals~\citep{levshina_tokenbasedtypology_2019}. Another justification could also be that these are languages with a very different "default" word order to English; this would further motivate~\citepos{ding_selfattentioncrosslingual_2020} use of cross-lingually reordered position markers.

\begin{figure}[t]
    \centering
        \includegraphics[width=0.5\textwidth]{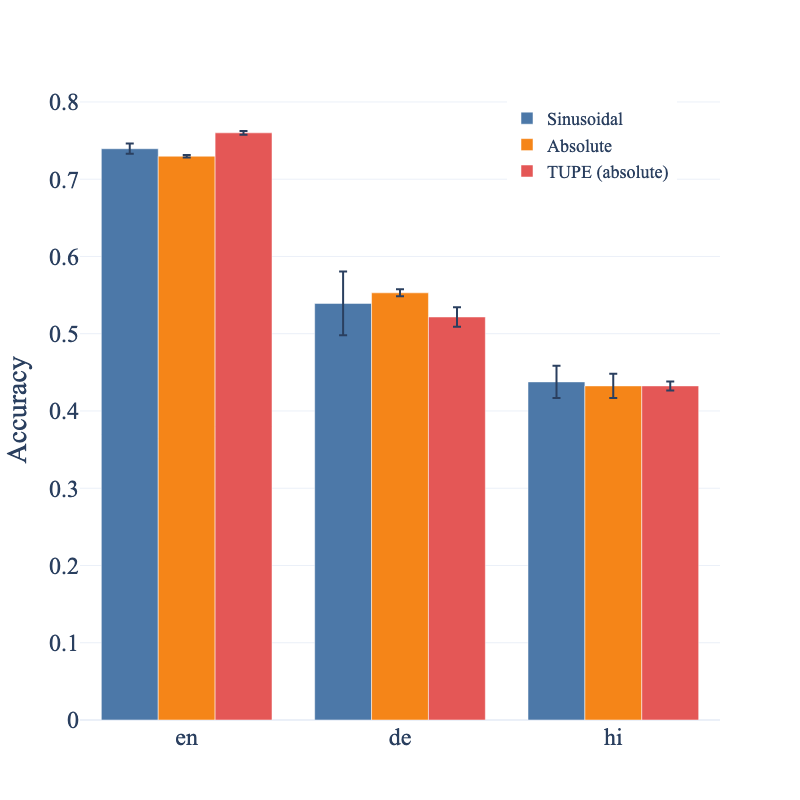}
    \caption{Real-world results on XNLI. Models were pretrained on a large text corpus and finetuned on English MultiNLI.}
    \label{fig:my_label}
\end{figure}

\paragraph{Real-world results}
While we conduct most of our analyses on our toy models, we also ran a series of experiments to verify that our results would hold with larger models. As such, we pre-trained full size BERT models (base, not large) for two epochs, on a corpus consisting of 8.5\textsc{m}, 9.3\textsc{m} and 800k sentences in English, German and Hindi respectively. We then fine-tuned these models for three epochs on (English) MultiNLI~\citep{williams_broadcoveragechallenge_2018}, and evaluated on held-out XNLI test sets for our three languages~\citep{conneau_xnlievaluating_2018}; the process took approximately 4 days per model, on a single V100 GPU. We trained two models (seeds 0 and 42) per method, for three different positional encoding methods: a) absolute positional encodings, as these are used in the original BERT, b) sinusoidal encodings, as these were the original transformer encodings, and c) TUPE (absolute), as the most recent innovation. Our real-world results appear to validate our toy experiments: performance on English, the language the model was fine-tuned on, is highest with TUPE, while cross-lingual transfer suffers, both on German and to a lesser extent on Hindi. 
\section{Analyses}

In an attempt to explain the significantly improved cross-lingual performance of absolute positional encodings, we tried to examine precisely what sort of encoding was being learnt. Part of the original motivation behind sinusoidal encodings was that they would allow for {\bf compositionality}; 
for any fixed offset $k$, there exists a linear transformation from $p_{pos}$ to $p_{pos+k}$, 
making it easier to learn to attend to relative offsets; the proof of this is in Appendix~\ref{sec:app_proof}.\footnote{\citet{vaswani_attention_2017} do not explicitly mention compositionality, but only generalization across positions for fixed offsets. Positional disentanglement is the flipside of compositionality, however \cite{chaabouni-etal-2020-compositionality}.}

\begin{figure}[t]
    \centering
    \includegraphics[width=0.5\textwidth]{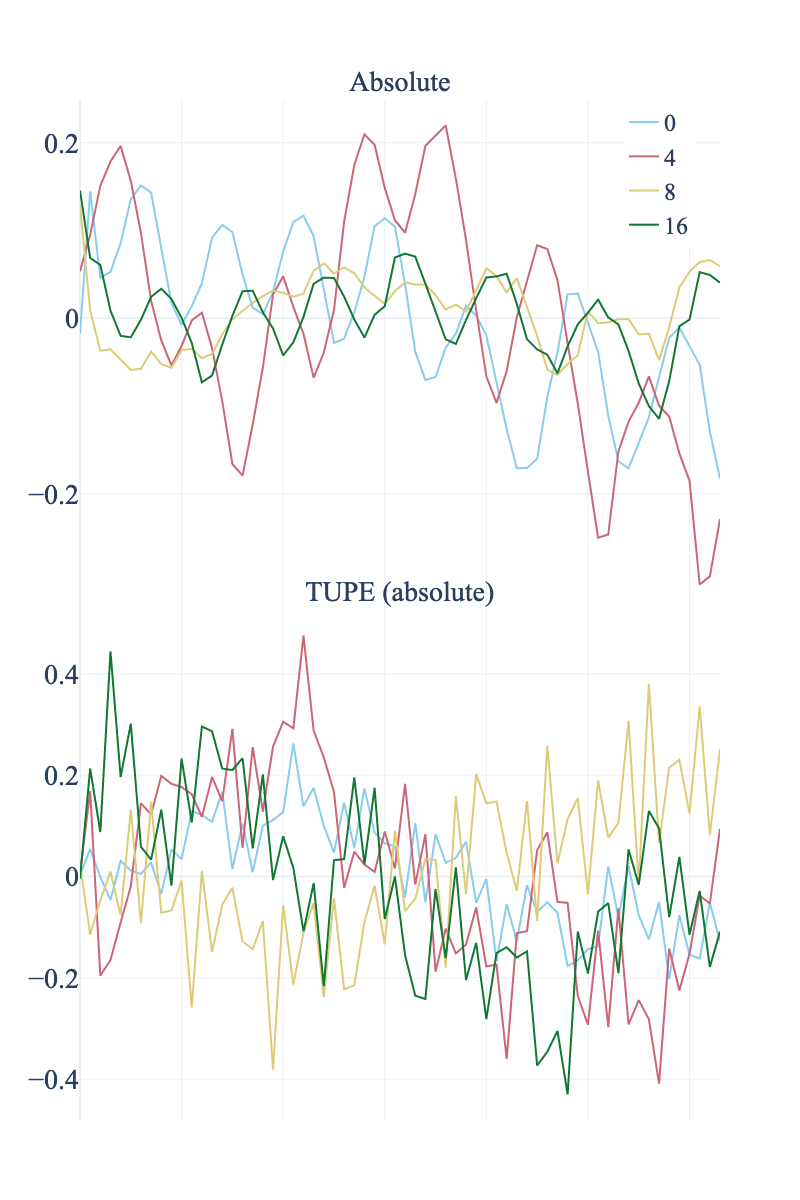}
    \caption{Dimensions 0, 4, 8 and 16 of learnt absolute and TUPE positional encodings over 32 positions for \enxx{fi}, seed 0.}
    \label{fig:abs_sin}
\end{figure}

We examined our absolute positional encodings to see whether or not they were being induced to learn some specific function. Figure~\ref{fig:abs_sin} plots 4 dimensions of absolute and TUPE(a) positional encoding, for the \biling{fi} model; each line represents a specific dimension of the encoding vectors generated for positions 0 to 31.  Interestingly, it appears that absolute representations converge to waveforms that represent sinusoids somewhat, while neither of the untied experiments do so (cf. Appendix~\ref{sec:app_plots}).

We hypothesize that absolute representations converge to waveforms because of increased pressure for compositionality, being trained on structurally different languages. To test this, we quantify the extent to which the absolute, relative and untied encodings are compositional in the sense that there is a linear transformation from $p_{pos}$ to $p_{pos + k}$ for different $k$. 

\begin{figure}[t]
    \centering
    \includegraphics[width=0.5\textwidth]{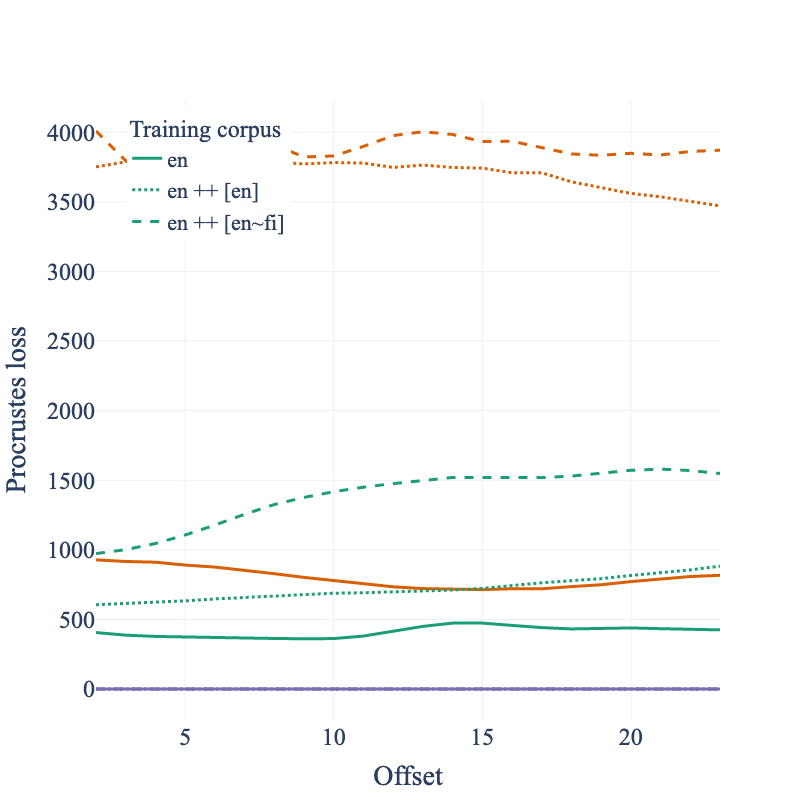}
    \caption{Procrustes loss for \textcolor{ForestGreen}{absolute} encodings and \textcolor{Orange}{TUPE} (seed 0); differences are statistically significant with $p < 0.001$ (Wilcoxon). The \textcolor{Violet}{sinusoidal} loss is $\approx$ 0.}
    \label{fig:procrustes}
\end{figure}
To this end, we use Procrustes analysis \cite{Stegmann:2002} to learn a linear transformation for each $k$, based on the representations of $p_{pos}$ and $p_{pos+k}$. Specifically, we apply \emph{orthogonal} Procrustes analyses~\citep{schonemann1966generalized}, which avoid scaling and translation. 

First, we minimise $\arg\min_\mathrm{T}||p_{pos}-\mathrm{T}p_{pos+k}||^2$. Next, we apply $\mathrm{T}$ to a different randomly selected $\mathit{pos'}$, i.e. we calculate $\mathcal{L} = ||p_{pos'}-\mathrm{T}p_{pos'+k}||^2$. The higher the final loss $\mathcal{L}$, the less our encodings facilitate compositionality. In order to make learning $\mathrm{T}$ simpler, rather than selecting representations for single positions $pos$ and $pos'$, we select chunks of arbitrary size $C$, and stack their positions into a matrix. Note that for sinusoidal representations, the loss is close to zero regardless of span. 

The losses are plotted over a range of offsets for both absolute representations and for TUPE(a), in Figure~\ref{fig:procrustes}; we include a control model trained on a monolingual corpus. Losses are averaged over 125 runs per offset, with random values of $\mathit{pos}$, $\mathit{pos'}$ and $C$. While both forms of representation appear to be similar (and relatively non-sinusoidal) when trained on the monolingual corpus, introducing bilingualism leads to a clear difference between the two: absolute positional representations tend to be a lot closer to sinusoidal representations than untied ones do. 
Note, also, that this gap is clearest for the (simpler) \orig{} experiment -- this is unsurprising, as \orig{} is still \emph{perceived} as bilingual due to the shifted vocabulary. The structural similarity between the two, however, makes it easier to build compositional representations by relying on offsets, as the model only needs to learn to represent one language, structurally speaking. We observe a similar gap when comparing pretrained BERT models: \textsf{bert-base-multilingual-cased} exhibits more sinusoidal representations over a range of offsets, when compared to \textsf{bert-base-cased}, although the gap is narrower than with our toy models. 
\begin{figure}[h]
    \centering
    \includegraphics[width=0.5\textwidth]{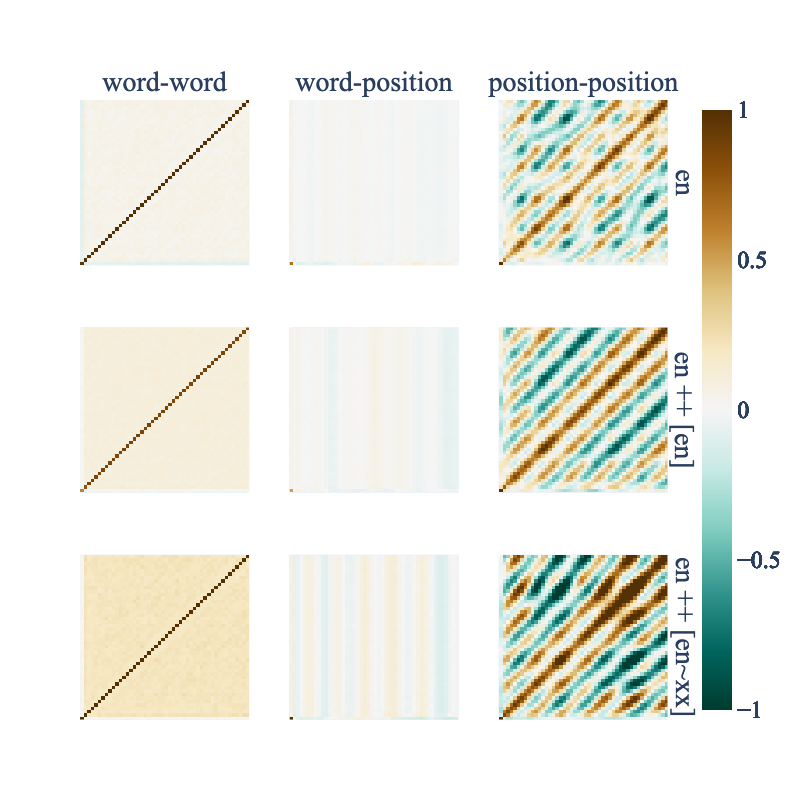}
    \caption{Word-position correlations for our Finnish-reordered model with random seed 0; words on the y- and positions on the x-axis.}
    \label{fig:heatmap}
\end{figure}

\paragraph{Correlations in multilingual settings}
A key motivation for eliminating word-position correlations, presented in~\citep{keRethinkingPositionalEncoding2020}, is the fact that these correlations are effectively zero, leading to no additional information for the model. Figure~\ref{fig:heatmap} captures word-position correlations from three of our trained models (with an additional model trained on a purely monolingual corpus); note that while these correlations are very close to zero for monolingual corpora, there is a visible "banding" phenomenon in the multilingual corpora, that only grows stronger when a different grammar is sampled. A similar banding phenomenon is visible when we compare multilingual and monolingual pre-trained BERT models (Appendix~\ref{sec:app_plots}), albeit with reduced magnitude. We hypothesize that the pressure for compositionality induces these correlations. 


\begin{figure}
    \centering
    \includegraphics[width=0.5\textwidth]{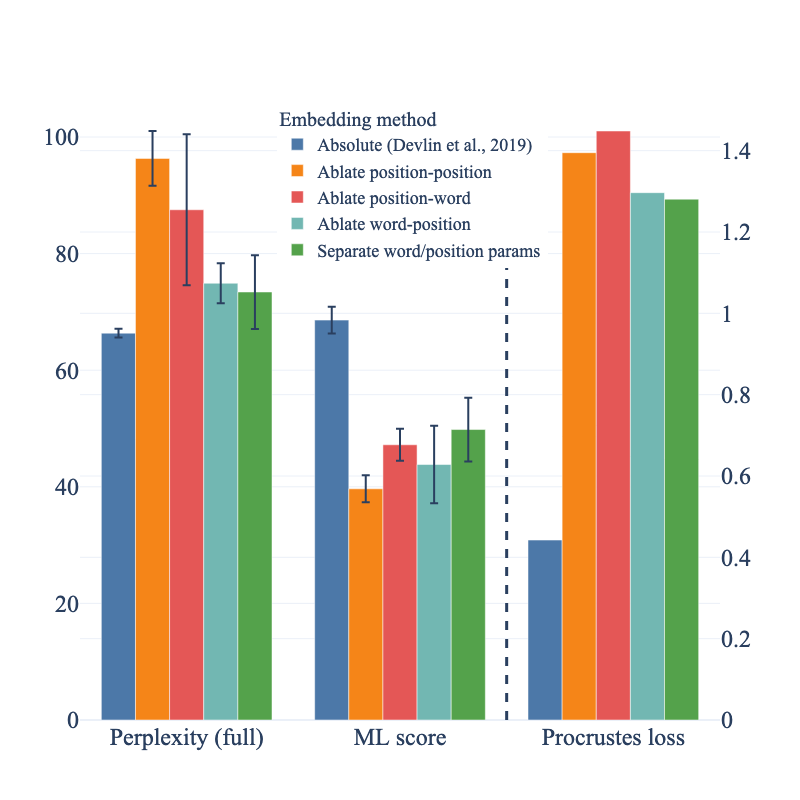}
    \caption{Ablation experiments, averaged over languages (for perplexity and ML score). Procrustes losses calculated as in \S5, for the \enxx{fi} model (seed 0).}
    \label{fig:ablate}
\end{figure}

\paragraph{Ablation studies}
Finally, we ran a series of ablation experiments on absolute positional encodings to support the above analysis. Three of the experiments involved removing position-position correlations, position-word correlations, word-position correlations, and a fourth involved using separate parameters for word and position attention. Results are presented in Figure~\ref{fig:ablate}; we also include the median Procrustes loss. 
We note that the removal of both position-word correlations and word-position correlations has an effect on both perplexity and ML score. Interestingly, removing word-position correlations ($(p_iW^Q)(w_jW^K)^\top{}$) does not have the same effect as the inverse does: perplexity is lower than with position-word correlations removed, but so is the ML score, indicating a difference between the role played by position as a key, and as a query. 

\paragraph{On relative representations}
Given our previous assumptions about offsets aiding compositionality, why, then, do our relative representations - that explicitly calculate offsets - perform poorly in multilingual settings? We speculate that the reason relative encodings appear to hurt multilingual compression is that offset-specific bias terms sparsify the learning signal for (and thereby hinder the alignment of) disjoint vocabularies. In compensating for this, relative positional encodings sacrifice their compositionality. Relative representations aid compositionality by directly providing a bias term derived from the distance between a word pair. As shown above, absolute representations learn similar biases; however, being actively forced to learn such biases could encourage models to jointly learn alignment and compositionality. 

Further, offset representations are also effectively "hard", i.e. derived from the hard distance between the two tokens. The interaction between $w_i$ and $w_j$ is not wholly mediated by the distance $i - j$, however, this correlation is forced by the product term $(x_iW^Q)(a_{ij})^\top{}$. The term $(x_iW^Q)(p_jW^K)^\top{}$, on the other hand, could effectively attend to multiple offsets. $p_jW^K$ is fixed for position $j$; given the sinusoidal nature of $p$, the product term could induce a "soft" positional representation with subspaces attending to different offsets\footnote{Indeed, we find that $p_jW^k$ is less invariant to Procrustes transformation than $p_j$ is.}; the relevant offset mix could then be indexed into by $x_iW^Q$.

\section{Discussion}

The main contribution of our work is practical, namely showing that findings about positional encodings in the context of monolingual language models do not apply straightforwardly to multilingual language models. In answering why sinusoidal embeddings are superior to more recent alternatives in the multilingual setting, we also found the compositionality of positional encodings to be predictive of multilingual compression in such models. While relative positional encodings seem designed for compositionality, they prevent efficient alignment of multilingual vocabularies. 

\citet{sinha_maskedlanguage_2021} show that word order matters little for monolingual language model pretraining, and that pretrained language models seem to rely mostly on higher-order word
co-occurrence statistics. 
Our work shows that this finding does not generalize to pretraining multilingual language models. In the multilingual setting, 
word order clearly matters, as also shown in previous work~\citep{keRethinkingPositionalEncoding2020,dufterIdentifyingNecessaryElements2021}, and compositional positional encodings seem to facilitate effective multilingual compression. 
This aligns with the observation that syntactic reordering 
à la \citet{ding_selfattentioncrosslingual_2020} is in some cases an effective way to encourage compositional cross-lingual representations.

In general, our results illustrate how methods developed for monolingual language models should not be blindly adopted when training multilingual models, which potentially require different architectures. Conversely, we would encourage future work on new positional encoding mechanisms for non-autoregressive models to also evaluate these mechanisms in multilingual settings. 


\section{Conclusion}

Through a series of synthetic and real experiments with training multilingual language models, we showed that a) sinusoidal positional encodings perform better in multilingual settings than more recent alternatives (that have been shown to perform better in monolingual settings); b) this is likely because of an increased pressure for compositionality. We devised a method for quantifying the compositionality of positional encodings, and strengthened our results by also considering word-position correlations and ablation studies. 

\bibliography{anthology,custom}
\bibliographystyle{acl_natbib}

\appendix

\clearpage

\section{Proof of linear transformability}
\label{sec:app_proof}

Let \[\begin{bmatrix}\sin(\omega{}t) \\ \cos(\omega{}t)\end{bmatrix}\] represent a sine/cosine pair, characterised by position $t$. Let \[\mathrm{R}_k = \begin{bmatrix}\cos(\omega{}k) & \sin(\omega{}k) \\ -\sin(\omega{}k) & \cos(\omega{}k)\end{bmatrix}\] be a rotation matrix for angle $\omega{}k$. We then have:

\begin{align*}
    R\begin{bmatrix}\sin(\omega{}t) \\ \cos(\omega{}t)\end{bmatrix} &= \begin{bmatrix}\cos(\omega{}k) & \sin(\omega{}k) \\ -\sin(\omega{}k) & \cos(\omega{}k)\end{bmatrix}\cdot\begin{bmatrix}\sin(\omega{}t) \\ \cos(\omega{}t)\end{bmatrix}\\
    &= \begin{bmatrix}\sin(\omega{}k)\cos(\omega{}t) + \cos(\omega{}k)\sin(\omega{}t)\\
    \cos(\omega{}k)\cos(\omega{}t) - \sin(\omega{}k)\sin\omega{}t)\end{bmatrix}\\
    &= \begin{bmatrix}\sin(\omega{}(t + k)) \\ \cos(\omega{}(t + k))\end{bmatrix}
\end{align*}

implying that for a fixed frequency $\omega$, there exists a rotation matrix $\mathrm{R}_k$ that can induce a rotational offset of $k$.
\section{Additional plots}
\label{sec:app_plots}
\begin{figure*}
    \centering
    \includegraphics[width=\textwidth]{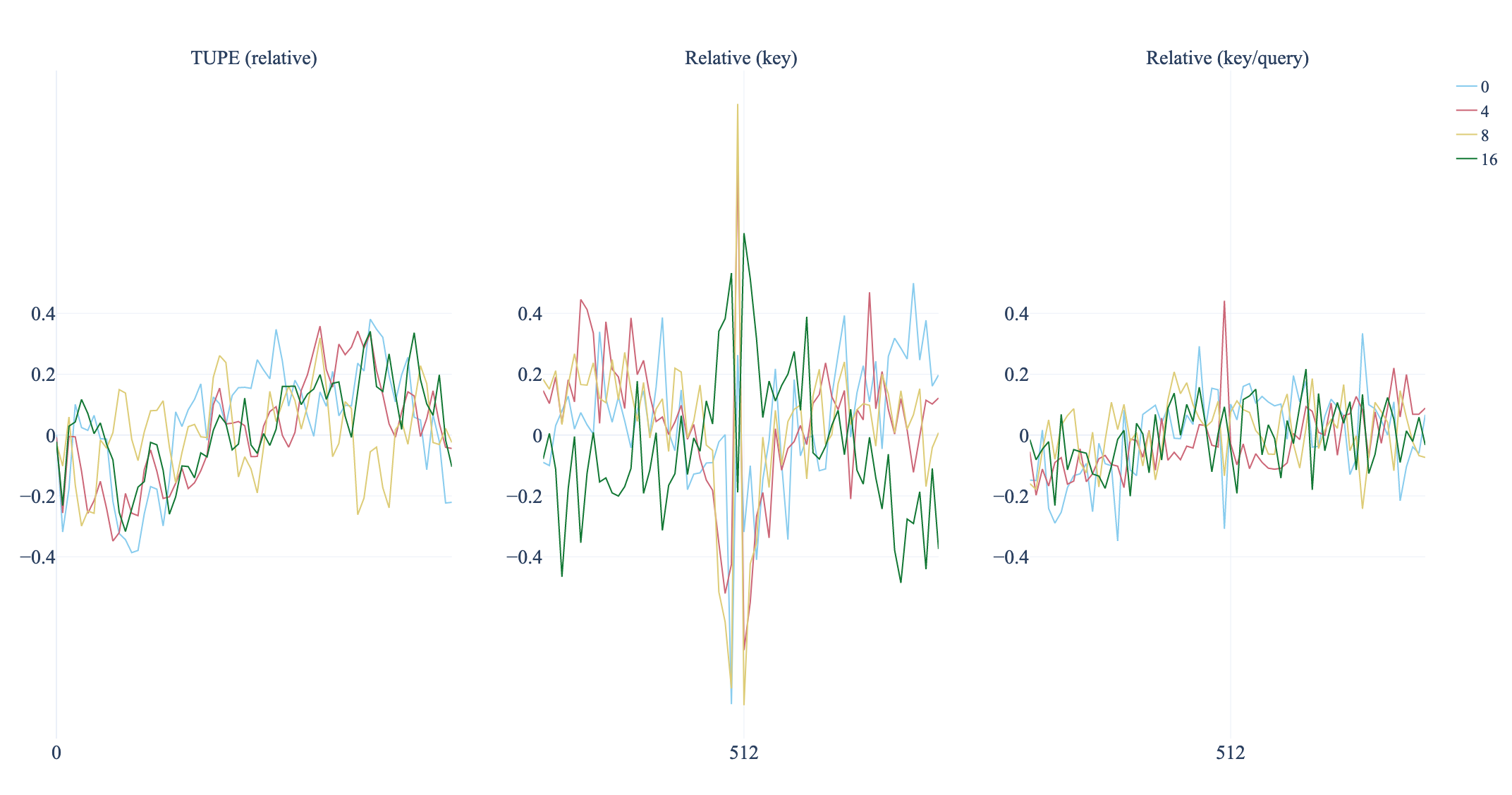}
    \caption{Four neurons over 32 for TUPE (relative); the same neurons for 32 offsets centred on 512 for the other relative models.}
    \label{fig:neurons_all}
\end{figure*}

\begin{figure}
    \centering
    \includegraphics[width=0.5\textwidth]{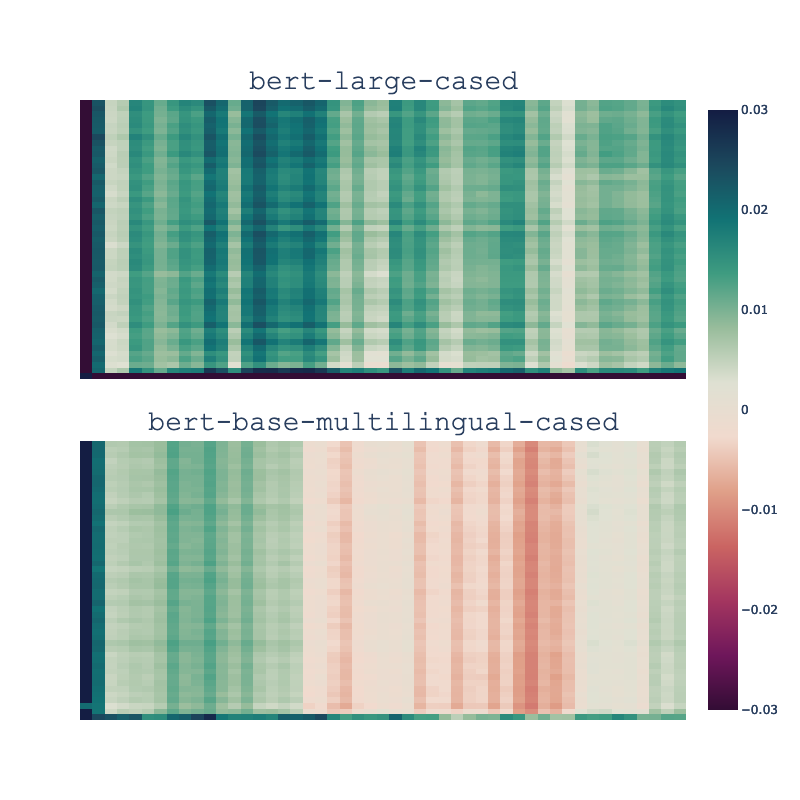}
    \caption{Word-position correlations for pretrained BERT models}
    \label{fig:bert}
\end{figure}
\clearpage
\section{Full results}
\label{sec:app_full}
\begin{table*}[t]
    \centering
    \resizebox{\textwidth}{!}{\begin{tabular}{cc|ccccccc}
    \multirow{2}{*}{\textbf{Wiki/CC}} & \multirow{2}{*}{Embedding} & \multicolumn{2}{c}{Perplexity} & \multicolumn{2}{c}{Retrieval} & \multicolumn{2}{c}{Translation} & \multirow{2}{*}{ML score} \\
    & & Full & L1 & 0 & 8 & 0 & 8 & \\
    \midrule
    \multirow{6}{*}{Arabic} &Sinusoidal &67.52 (1.72) &64.16 (2.14) &47.90 (25.39) &98.09 (0.90) &71.75 (7.39) &62.52 (0.09) &70.36 (5.73) \\
    &Absolute &68.54 (2.01) &65.02 (2.83) &44.84 (8.00) &95.14 (0.98) &70.79 (3.62) &46.74 (10.01) &63.67 (3.37) \\
    &TUPE (a) &62.09 (0.57) &59.01 (0.45) &7.88 (1.63) &75.17 (7.47) &58.06 (4.65) &32.13 (3.28) &40.15 (4.45) \\
    &TUPE (r) &61.22 (0.50) &57.81 (1.05) &15.12 (4.39) &83.44 (13.58) &64.94 (9.55) &29.03 (10.33) &46.22 (10.09) \\
    &Relative (k) &58.08 (0.98) &53.92 (0.89) &12.61 (5.00) &74.14 (15.52) &63.27 (13.04) &9.37 (3.85) &38.56 (9.24) \\
    &Relative (k/q) &55.96 (1.46) &51.98 (1.02) &20.95 (8.34) &89.53 (8.39) &73.01 (5.47) &35.83 (20.20) &55.23 (12.46) \\
    \midrule
    \multirow{6}{*}{German} &Sinusoidal &67.34 (1.52) &64.74 (1.21) &47.12 (25.19) &98.94 (0.22) &71.53 (14.37) &61.06 (10.25) &70.95 (3.74) \\
    &Absolute &65.28 (1.69) &62.67 (2.08) &63.68 (2.44) &98.87 (0.06) &80.21 (0.60) &36.34 (24.86) &67.01 (8.44) \\
    &TUPE (a) &61.29 (0.70) &58.70 (0.60) &12.11 (1.29) &92.87 (2.50) &71.39 (3.10) &48.28 (3.36) &55.15 (2.09) \\
    &TUPE (r) &59.42 (0.77) &56.90 (1.29) &18.27 (3.33) &97.40 (1.13) &77.86 (2.10) &43.72 (13.84) &60.74 (5.49) \\
    &Relative (k) &56.00 (0.22) &52.29 (1.23) &23.00 (5.06) &95.70 (3.57) &79.66 (1.94) &26.08 (17.72) &55.67 (8.05) \\
    &Relative (k/q) &55.45 (0.73) &51.77 (1.49) &22.67 (7.54) &90.80 (10.64) &76.50 (7.72) &39.51 (18.52) &56.83 (11.69) \\
    \midrule
    \multirow{6}{*}{Basque} &Sinusoidal &68.44 (0.88) &64.15 (1.30) &48.65 (24.10) &97.01 (0.70) &70.63 (7.73) &61.65 (2.26) &70.16 (5.46) \\
    &Absolute &69.46 (3.29) &65.36 (2.70) &45.42 (5.55) &91.64 (3.68) &69.19 (3.87) &36.91 (6.75) &58.51 (3.31) \\
    &TUPE (a) &63.11 (0.36) &59.23 (1.35) &6.38 (2.17) &71.92 (12.96) &57.18 (7.90) &28.21 (16.17) &39.13 (10.48) \\
    &TUPE (r) &60.82 (0.86) &56.61 (1.77) &10.85 (2.47) &77.57 (12.22) &64.37 (6.83) &33.41 (12.54) &43.95 (8.01) \\
    &Relative (k) &60.03 (0.57) &54.70 (0.53) &9.49 (4.37) &61.06 (20.25) &56.12 (14.21) &21.28 (15.21) &36.27 (12.29) \\
    &Relative (k/q) &57.66 (0.96) &52.51 (1.41) &11.74 (6.91) &63.13 (32.58) &55.84 (22.94) &33.70 (18.55) &41.27 (20.79) \\
    \midrule
    \multirow{6}{*}{Finnish} &Sinusoidal &67.28 (1.54) &64.20 (1.88) &51.99 (23.78) &98.90 (0.25) &76.33 (5.69) &69.72 (1.44) &74.51 (5.15) \\
    &Absolute &67.14 (2.20) &64.04 (2.76) &51.91 (11.81) &97.90 (1.13) &77.54 (0.67) &63.53 (7.04) &72.83 (1.87) \\
    &TUPE (a) &61.53 (1.21) &58.06 (0.95) &11.45 (1.36) &91.69 (4.67) &71.60 (3.69) &34.34 (16.61) &50.91 (6.36) \\
    &TUPE (r) &61.07 (1.22) &57.54 (1.07) &15.90 (2.15) &94.00 (3.82) &75.84 (4.08) &47.95 (20.72) &58.29 (7.04) \\
    &Relative (k) &57.05 (1.14) &53.13 (0.43) &22.31 (3.20) &96.01 (0.70) &78.96 (0.68) &20.75 (7.37) &53.76 (0.78) \\
    &Relative (k/q) &55.37 (0.66) &51.29 (0.27) &22.97 (0.86) &91.87 (5.67) &79.77 (0.97) &22.81 (25.52) &54.63 (9.17) \\
    \midrule
    \multirow{6}{*}{French} &Sinusoidal &63.92 (0.41) &61.82 (0.87) &54.10 (22.52) &99.36 (0.13) &77.65 (4.95) &70.28 (1.22) &75.76 (4.25) \\
    &Absolute &64.76 (1.10) &62.68 (1.37) &59.12 (13.35) &99.14 (0.37) &79.57 (0.63) &68.90 (2.75) &76.28 (1.93) \\
    &TUPE (a) &59.94 (0.79) &58.60 (0.71) &8.67 (2.15) &90.25 (7.76) &67.86 (7.47) &36.15 (7.99) &50.87 (6.57) \\
    &TUPE (r) &58.28 (0.78) &56.14 (1.37) &19.36 (3.62) &96.64 (2.93) &78.50 (1.72) &51.63 (18.73) &63.50 (7.99) \\
    &Relative (k) &55.18 (0.25) &52.96 (0.44) &18.84 (2.14) &91.12 (8.32) &73.30 (7.86) &37.50 (17.77) &56.07 (8.40) \\
    &Relative (k/q) &54.35 (0.23) &52.10 (1.02) &19.01 (5.15) &90.09 (9.29) &72.22 (7.48) &26.51 (22.43) &51.88 (10.57) \\
    \midrule
    \multirow{6}{*}{Hindi} &Sinusoidal &67.46 (0.27) &65.01 (0.32) &37.94 (20.74) &88.30 (4.64) &63.26 (13.46) &47.88 (6.96) &60.76 (4.85) \\
    &Absolute &63.75 (1.05) &61.38 (0.97) &47.98 (0.66) &95.58 (2.19) &76.18 (1.95) &56.57 (12.48) &67.28 (4.54) \\
    &TUPE (a) &61.70 (1.10) &59.57 (0.48) &6.34 (1.56) &74.79 (20.15) &58.50 (12.29) &28.73 (6.46) &40.11 (9.49) \\
    &TUPE (r) &59.57 (1.24) &57.64 (1.75) &12.67 (6.57) &72.64 (24.01) &61.35 (17.78) &23.49 (10.80) &39.66 (14.67) \\
    &Relative (k) &55.50 (0.60) &52.81 (0.87) &19.72 (4.83) &90.11 (9.87) &74.06 (4.06) &26.71 (13.65) &52.39 (9.33) \\
    &Relative (k/q) &54.51 (0.21) &52.04 (0.74) &23.45 (6.88) &93.27 (6.35) &75.60 (3.41) &32.95 (21.46) &56.72 (10.94) \\
    \midrule
    \multirow{6}{*}{Swedish} &Sinusoidal &64.62 (0.86) &62.32 (0.73) &57.11 (21.45) &99.32 (0.24) &79.89 (2.06) &71.82 (7.63) &77.39 (5.05) \\
    &Absolute &65.54 (1.86) &62.95 (2.48) &53.52 (4.43) &97.45 (0.79) &78.85 (1.69) &66.33 (3.09) &74.58 (2.28) \\
    &TUPE (a) &60.16 (0.99) &58.23 (1.49) &14.47 (3.45) &96.36 (2.49) &76.62 (3.05) &48.12 (19.76) &60.16 (7.98) \\
    &TUPE (r) &58.29 (0.17) &56.12 (0.72) &21.57 (3.06) &97.80 (1.40) &79.05 (3.18) &54.56 (2.06) &64.86 (1.83) \\
    &Relative (k) &54.94 (0.61) &52.28 (0.68) &28.17 (3.30) &98.52 (0.68) &82.81 (1.08) &56.60 (9.17) &68.09 (5.04) \\
    &Relative (k/q) &53.66 (0.22) &51.16 (1.11) &27.87 (4.86) &97.63 (2.04) &82.11 (2.22) &49.65 (17.66) &65.89 (8.57) \\
    \bottomrule
    \end{tabular}}
    \label{tab:full_results_wiki}
\end{table*}

\begin{table*}[b]
    \centering
    \resizebox{\textwidth}{!}{\begin{tabular}{cc|ccccccc}
    \multirow{2}{*}{\textbf{Bible}} & \multirow{2}{*}{Embedding} & \multicolumn{2}{c}{Perplexity} & \multicolumn{2}{c}{Retrieval} & \multicolumn{2}{c}{Translation} & \multirow{2}{*}{ML score} \\
    & & Full & L1 & 0 & 8 & 0 & 8 & \\
    \midrule
    \multirow{6}{*}{Arabic} &Sinusoidal &228.44 (5.29) &193.76 (3.88) &2.30 (0.35) &32.79 (4.29) &24.54 (5.24) &13.20 (3.62) &18.21 (3.09) \\
    &Absolute &236.83 (8.40) &206.44 (10.04) &4.18 (0.20) &22.72 (5.92) &24.18 (3.04) &5.75 (1.87) &14.21 (2.67) \\
    &TUPE (a) &251.32 (5.55) &209.54 (9.00) &0.72 (0.36) &10.18 (6.12) &9.91 (5.73) &3.44 (2.47) &6.07 (3.67) \\
    &TUPE (r) &228.09 (18.36) &193.67 (7.37) &1.30 (0.22) &21.49 (6.68) &23.25 (3.57) &4.19 (1.11) &12.56 (2.28) \\
    &Relative (k) &206.77 (4.36) &177.43 (3.93) &0.84 (0.08) &8.79 (1.82) &11.85 (1.24) &3.24 (0.64) &6.18 (0.73) \\
    &Relative (k/q) &195.47 (5.84) &166.20 (2.53) &1.44 (0.60) &15.87 (8.02) &19.30 (7.44) &4.82 (2.64) &10.36 (4.52) \\
    \midrule
    \multirow{6}{*}{German} &Sinusoidal &214.19 (8.29) &185.11 (3.92) &6.80 (1.86) &76.04 (3.25) &64.60 (2.83) &37.76 (5.90) &46.30 (2.53) \\
    &Absolute &230.05 (7.75) &205.46 (12.23) &9.81 (2.28) &63.69 (12.37) &51.91 (5.98) &17.89 (5.89) &35.83 (6.55) \\
    &TUPE (a) &230.92 (6.27) &203.84 (2.26) &1.75 (0.39) &33.91 (2.86) &36.33 (6.21) &9.59 (2.96) &20.40 (2.65) \\
    &TUPE (r) &211.44 (6.19) &196.39 (7.09) &2.52 (0.56) &48.23 (20.03) &46.61 (14.36) &14.43 (6.78) &27.95 (9.95) \\
    &Relative (k) &189.45 (4.10) &165.36 (2.59) &1.50 (0.48) &16.44 (7.67) &19.43 (8.06) &5.12 (2.08) &10.62 (4.56) \\
    &Relative (k/q) &191.06 (0.27) &168.70 (6.74) &2.33 (0.67) &33.21 (15.24) &36.76 (13.08) &11.69 (6.65) &21.00 (8.84) \\
    \midrule
    \multirow{6}{*}{Basque} &Sinusoidal &236.91 (9.77) &197.71 (11.45) &1.47 (0.37) &15.82 (2.59) &13.57 (3.44) &6.47 (2.72) &9.33 (2.27) \\
    &Absolute &248.87 (18.53) &212.26 (14.12) &4.90 (0.82) &24.07 (4.75) &17.72 (2.05) &6.00 (2.44) &13.17 (1.15) \\
    &TUPE (a) &265.28 (11.20) &220.59 (17.75) &0.42 (0.18) &2.98 (1.30) &4.57 (1.17) &1.21 (0.57) &2.29 (0.79) \\
    &TUPE (r) &239.52 (8.88) &196.16 (12.43) &1.04 (0.21) &11.23 (1.54) &14.64 (4.25) &3.29 (0.98) &7.55 (1.36) \\
    &Relative (k) &214.59 (5.79) &170.40 (2.68) &0.52 (0.14) &3.24 (1.20) &6.23 (1.05) &1.60 (0.67) &2.90 (0.62) \\
    &Relative (k/q) &206.33 (6.43) &166.16 (2.67) &0.49 (0.18) &3.15 (1.43) &5.96 (2.18) &1.37 (0.86) &2.74 (1.09) \\
    \midrule
    \multirow{6}{*}{Finnish} &Sinusoidal &215.28 (2.80) &181.92 (5.08) &4.34 (0.29) &64.24 (9.50) &56.92 (5.55) &28.02 (8.05) &38.38 (5.73) \\
    &Absolute &230.22 (12.83) &194.61 (4.17) &14.40 (3.69) &68.82 (5.06) &63.92 (5.37) &26.18 (8.15) &43.33 (5.03) \\
    &TUPE (a) &251.91 (4.75) &215.41 (10.58) &1.94 (0.56) &35.29 (11.75) &36.96 (8.56) &10.58 (6.28) &21.19 (6.73) \\
    &TUPE (r) &240.35 (7.43) &206.09 (14.62) &3.00 (0.40) &52.44 (7.90) &53.54 (5.29) &9.22 (5.12) &29.55 (3.15) \\
    &Relative (k) &202.40 (8.09) &170.90 (9.54) &1.77 (0.90) &23.66 (13.30) &30.67 (14.17) &11.30 (5.45) &16.85 (8.23) \\
    &Relative (k/q) &197.92 (5.13) &161.33 (8.55) &2.38 (0.70) &29.05 (3.31) &40.63 (10.06) &12.74 (5.02) &21.20 (4.70) \\
    \midrule
    \multirow{6}{*}{French} &Sinusoidal &202.44 (4.97) &179.19 (6.13) &7.36 (1.66) &82.96 (1.61) &74.46 (1.00) &44.95 (4.13) &52.43 (0.76) \\
    &Absolute &217.10 (11.25) &199.77 (10.51) &13.60 (1.64) &75.18 (7.83) &74.33 (5.80) &35.35 (9.98) &49.62 (6.21) \\
    &TUPE (a) &220.48 (9.85) &203.21 (12.12) &2.49 (0.35) &52.95 (15.51) &46.47 (10.33) &15.98 (5.82) &29.47 (7.80) \\
    &TUPE (r) &204.57 (8.86) &188.30 (14.18) &4.07 (0.72) &71.33 (10.07) &65.27 (6.85) &29.41 (8.71) &42.52 (6.50) \\
    &Relative (k) &181.31 (2.67) &161.53 (3.44) &2.36 (1.09) &36.83 (20.95) &33.84 (15.39) &12.17 (9.58) &21.30 (10.92) \\
    &Relative (k/q) &179.43 (9.50) &159.12 (8.66) &4.01 (0.93) &57.36 (10.80) &56.62 (9.44) &24.81 (10.53) &35.70 (6.91) \\
    \midrule
    \multirow{6}{*}{Hindi} &Sinusoidal &207.71 (3.54) &187.60 (4.39) &2.08 (0.53) &22.03 (8.05) &17.30 (5.85) &7.54 (2.13) &12.24 (4.13) \\
    &Absolute &221.81 (6.29) &197.47 (9.65) &4.62 (1.64) &29.81 (13.95) &23.44 (9.49) &6.00 (1.86) &15.97 (6.55) \\
    &TUPE (a) &235.64 (11.94) &216.64 (15.13) &0.70 (0.06) &5.63 (1.00) &9.15 (2.07) &2.31 (0.57) &4.45 (0.69) \\
    &TUPE (r) &210.78 (8.59) &189.10 (9.12) &0.83 (0.37) &12.64 (5.58) &13.41 (6.40) &3.26 (1.76) &7.54 (3.52) \\
    &Relative (k) &190.88 (4.72) &172.32 (7.12) &1.24 (0.43) &15.41 (6.60) &14.80 (6.91) &4.71 (2.33) &9.04 (4.06) \\
    &Relative (k/q) &190.68 (6.60) &168.92 (4.95) &1.06 (0.32) &11.42 (5.32) &13.01 (4.36) &3.59 (0.56) &7.27 (2.62) \\
    \midrule
    \multirow{6}{*}{Swedish} &Sinusoidal &205.42 (4.67) &180.45 (2.27) &9.41 (0.30) &84.72 (2.59) &78.21 (0.52) &55.40 (6.57) &56.94 (2.23) \\
    &Absolute &220.06 (10.32) &197.17 (5.13) &15.82 (0.56) &83.29 (0.69) &77.68 (1.00) &38.35 (4.80) &53.79 (1.34) \\
    &TUPE (a) &220.78 (8.49) &201.86 (6.14) &3.56 (0.64) &61.11 (12.33) &58.59 (11.18) &13.75 (3.21) &34.25 (6.03) \\
    &TUPE (r) &212.54 (3.02) &192.31 (9.32) &5.00 (1.46) &72.05 (11.82) &66.01 (9.39) &33.61 (10.73) &44.17 (7.93) \\
    &Relative (k) &184.79 (6.02) &165.70 (6.85) &5.56 (1.40) &72.94 (3.63) &69.32 (5.43) &14.18 (5.07) &40.50 (2.83) \\
    &Relative (k/q) &177.74 (6.85) &160.42 (5.99) &5.08 (0.24) &71.11 (5.47) &67.52 (3.27) &32.22 (12.51) &43.99 (1.05) \\
    \bottomrule
    \end{tabular}}
    \caption{Full results (mean/std. over three seeds)}
\end{table*}

\end{document}